\documentclass{article}
\usepackage{spconf, amsmath, amssymb, graphicx}
\usepackage{xltabular}
\usepackage{array}
\usepackage{array}
\newcolumntype{Y}{>{\hsize=0.2\hsize}X} 
\usepackage{tabularx}
\usepackage[dvipsnames]{xcolor}
\usepackage{booktabs}   
\usepackage{siunitx}   
\usepackage{dcolumn}
\usepackage{booktabs}
\usepackage{amsmath}
\usepackage{cancel}
\usepackage{tabularx}
\usepackage[table]{xcolor}
\usepackage{soul}
\usepackage{hyperref}
\usepackage{booktabs}
\usepackage{colortbl}
\usepackage{xcolor}

\usepackage{multirow}
\usepackage[normalem]{ulem}

\title{
Beyond Attention Scores: SVD-Based Vision Token Pruning \\ for Efficient Vision-Language Models}

\name{Yvon Apedo$^{\star }$ \qquad Martyna Poreba$^{\dagger}$ \qquad Michal Szczepanski$^{\dagger}$ \qquad Samia Bouchafa$^{\star}$}

\address{
$^{\star}$ Université Paris-Saclay, Univ Evry, IBISC, 91020, Evry-Courcouronnes, France  \quad \\
$^{\dagger}$ Université Paris-Saclay, CEA, List, F-91120, Palaiseau, France \quad
}

\begin{document}
%
\maketitle
\begin{abstract}
 Vision-Language Models (VLMs) have revolutionized multimodal learning by jointly processing visual and textual information. Yet, they face significant computational and memory challenges due to the large number of visual tokens processed during inference. Existing token pruning methods typically estimate token importance using local heuristics, such as attention scores, token norms, or similarity measures, which may fail to capture the global structure of visual representations and become less reliable under aggressive pruning. To address this limitation, we propose SVD-Prune, a training-free, plug-and-play token pruning method based on Singular Value Decomposition (SVD). Our method decomposes the vision token feature matrix into its principal components and uses statistical leverage scores to identify the tokens contributing most to the dominant spectral structure of the image. As a result, SVD-Prune preserves globally informative visual content while remaining independent of attention mechanisms and model-specific architectures. Extensive experiments on multiple VLM benchmarks demonstrate that SVD-Prune consistently outperforms existing pruning methods, maintaining strong multimodal reasoning performance even with only 32 or 16 visual tokens. Code: \url{https://github.com/anoncvlab/svdPrune}.


\end{abstract}
\begin{keywords}
Vision–Language Models, Token Pruning, Computational Efficiency, Singular value decomposition
\end{keywords}
\section{Introduction}
\label{sec:intro}

The rapid advancement of Large Language Models (LLMs) has driven remarkable progress in Vision-Language Models (VLMs), which integrate visual and textual modalities to enable sophisticated multimodal reasoning. Typical VLMs convert images into discrete vision tokens via a vision encoder and process these tokens sequentially alongside text through an LLM decoder. Through techniques such as modal alignment and instruction tuning, these models effectively adapt the strong perception and reasoning capabilities of LLMs to the visual domain. While this paradigm has yielded impressive results on cloud-scale infrastructure, deploying and continuously adapting VLMs on resource-constrained edge devices remains challenging due to the prohibitive memory and computational demands of backpropagation, where activation storage, particularly for vision tokens, often dominates the footprint. To alleviate these limitations, recent efforts have explored lightweight VLM architectures with reduced parameter counts and efficient visual token processing. For instance, LLaVAMini~\cite{zhang2025llavamini} employs a pre-fusion stage to jointly encode visual and textual tokens, followed by a compression module that reduces visual representations before they are processed by the language model.
However, these small-footprint models typically require extensive training or fine-tuning on large-scale multi modal datasets, limiting their accessibility and adaptability in resource-limited settings. An alternative, increasingly popular direction focuses on pruning vision tokens to mitigate redundancy in visual representations. Empirical studies reveal that vision tokens exhibit substantial redundancy, with the LLM decoder often attending more heavily to text tokens than to visual ones.
Consequently, numerous methods prioritize selecting salient vision tokens while discarding less informative ones. 
Despite recent advances in accelerating VLM inference, most existing token pruning approaches estimate token importance from local cues rather than the global structure of visual representations. In decoder-side methods, these cues often include attention scores, norms, or cross-modal similarities, which are susceptible to positional biases and attention-sink effects. These limitations often lead to inconsistent performance degradation, particularly at low token budgets or in complex scenes where preserving diverse semantic and spatial information is essential.

To overcome these challenges, we propose an SVD-based vision token pruning technique adapted from low-rank tensor decomposition. Our method decomposes the vision token feature matrix into principal components to capture its global spectral structure and computes leverage scores to estimate token importance. It then retains the smallest subset preserving a target fraction of spectral energy while maintaining spatial order for positional consistency.
Experimental results show that SVD-Prune achieves a strong efficiency--accuracy trade-off across a wide range of token budgets. In particular, the proposed method remains highly competitive in the low-token regime while significantly reducing computational cost and inference latency.

\section{RELATED WORK}
\label{sec:related_work}

VLMs encode images into substantially more tokens than text, making visual representations the primary computational bottleneck. This imbalance stems from spatial redundancy and semantic sparsity in visual data, leading to increased  memory usage, computational cost, and inference latency. As a response, recent approaches have investigated token pruning mechanisms that leverage attention-derived importance scores to reduce visual redundancy. Depending on the pruning stage, these methods operate either within the vision encoder~\cite{bolya2023tome, zhang2024cls, yang2025visionzip, song2025trim, arif2025hired, alvar2025divprune}, or during multimodal decoding~\cite{chen2024fastv,ye2025fitprune, zhang2025sparsevlm, wang2025PDrop}. Adaptive, multi-stage pruning strategies across the vision–language pipeline has been also explored~\cite{huang2024ivtp}.

Decoder-side methods primarily differ in how pruning decisions are controlled across layers. FastV~\cite{chen2024fastv} adopts fixed token budgets, FitPrune~\cite{ye2025fitprune} relies on offline, recipe-driven pruning schedules, while SparseVLM~\cite{zhang2025sparsevlm} introduces adaptive, text-aware pruning with token recycling. PyramidDrop~\cite{wang2025PDrop} further enforces a depth-aware strategy by progressively increasing the pruning ratio across decoder layers.
Encoder-side pruning strategies differ in whether token reduction is embedded directly into vision backbone, as in ToMe~\cite{bolya2023tome} or applied as a post-processing step on the encoder outputs~\cite{zhang2024cls, yang2025visionzip, song2025trim, arif2025hired, alvar2025divprune}, thereby preserving the original visual computation. HiRED~\cite{arif2025hired} and FasterVLM~\cite{zhang2024cls} apply attention-guided, one-shot token discarding based on \texttt{[CLS]} attention under controlled token budgets. VisionZip~\cite{yang2025visionzip} augments token selection with similarity-based merging to retain contextual information, while TRIM~\cite{song2025trim} introduces text-aware token selection via CLIP similarity and complements pruning with an aggregated token.

In practice, most token pruning methods are evaluated under moderate post-pruning token budgets, typically retaining 64-128 vision tokens, where they achieve a favorable trade-off between efficiency and accuracy. Performance in low-token regime (e.g., 32 or 16 tokens) remains underexplored in existing VLM pruning literature.

\section{METHODOLOGY}
\label{sec:methodology}

\begin{figure*}[!t]
    \centering
    \includegraphics[width=1.0\linewidth]{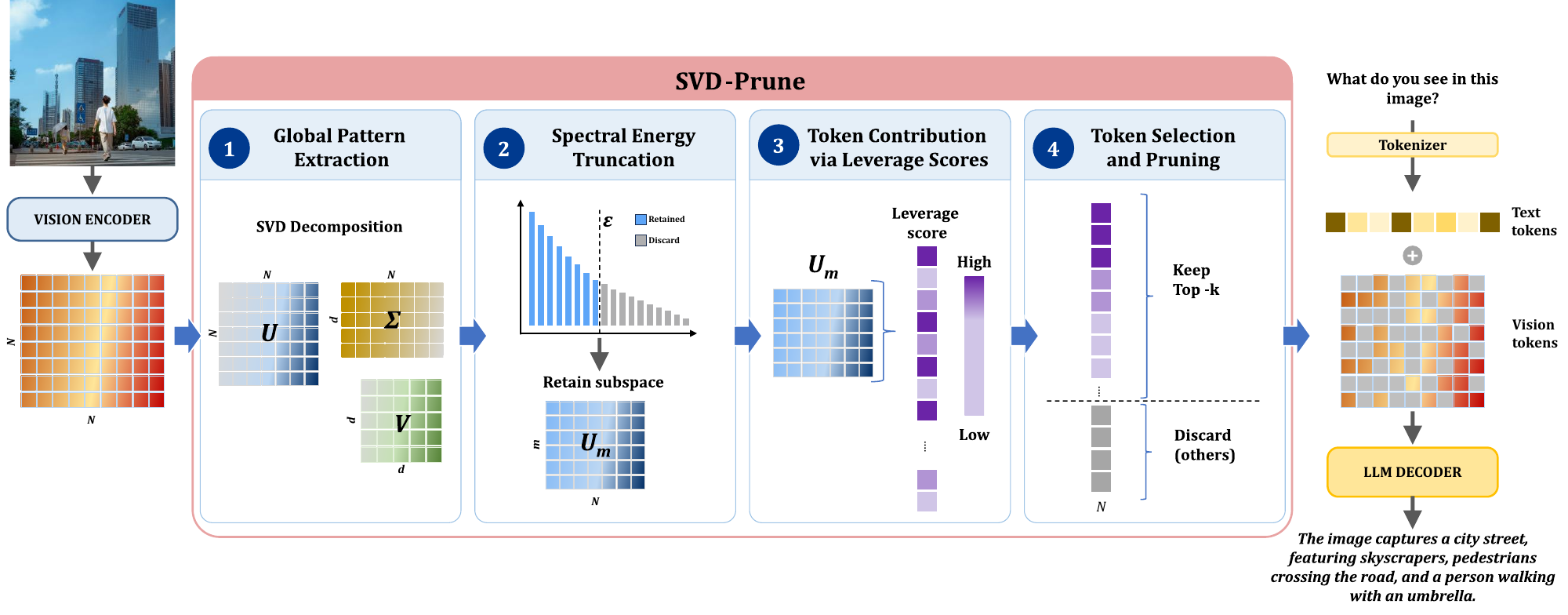}
    \caption{Overview of SVD-Prune. The method performs outside-encoder vision token pruning by applying a global SVD decomposition to vision encoder outputs, estimating token importance via leverage scores, and selecting a compact subset of informative vision tokens before multimodal decoding. $N$ denotes the number of tokens, $d$ the hidden dimension, and pruned tokens are highlighted in grey. }
    \label{fig:arch_svd}
\end{figure*}




We propose SVD-Prune, a training-free and attention
independent vision token pruning method that selects informative tokens based on their global representational structure. We reinterpret vision token pruning as a low-rank approximation problem, where the objective is to preserve the dominant semantic subspace using only a compact subset of tokens. As illustrated in Fig.~\ref{fig:arch_svd}, SVD-Prune operates outside the vision encoder on fully contextualized visual representations, preserving the backbone and enabling a plug-and-play design. Our approach proceeds in four stages. First, it performs a Singular Value Decomposition (SVD) to capture global visual patterns. Next, it truncates the decomposition to retain only dominant spectral energy components. Then, it estimates each token’s contribution using leverage scores. Finally, it selects and prunes tokens based on their cumulative importance. 

\subsection{Global Pattern Extraction}
\label{ssec:extraction}

Given an input image processed by the vision encoder, we obtain a feature matrix $\mathbf{F} \in \mathbb{R}^{N \times d}$, where $N$ is the number of vision tokens and $d$ is the hidden dimension. To capture the global structure across all tokens, we perform a SVD on the entire feature matrix :
\begin{equation}
\mathbf{F} = \mathbf{U} \boldsymbol{\Sigma} \mathbf{V}^\top,
\label{eq:decomposition}
\end{equation}
where the matrix $\mathbf{U} \in \mathbb{R}^{N \times N}$ contains the left singular vectors, whose columns represent the principal directions in token space, i.e., the most important linear combinations of tokens that capture the dominant patterns across the entire feature set. $\boldsymbol{\Sigma} \in \mathbb{R}^{N \times d}$  is the diagonal matrix of descending singular values $s_1 \geq s_2 \geq \cdots \geq s_d > 0$, which quantify the amount of spectral energy (or "strength") explained by each principal direction, sorted from most to least significant. And the matrix $\mathbf{V}^\top  \in \mathbb{R}^{d \times d}$ contains the right singular vectors (transposed), whose rows define the principal directions in feature space, \textit{i.e.} the key linear combinations of the original dimensions that form each important pattern. Together, these components allow us to identify and prioritize the global spectral energy structure of the vision tokens. Unlike attention-based metrics relying on local pairwise interactions, SVD provides a global low-rank view of token redundancy and informativeness.

\subsection{Spectral Energy Truncation
}
\label{ssec:var_pattern}
After decomposition, we identify the dominant spectral components by analyzing the energy associated with the singular values. Specifically, we compute the normalized spectral energy of each component as: 
\begin{equation}
e_i = \frac{s_i^2}{\sum_{j=1}^{\min(N,d)} s_j^2},
\label{eq:variance}
\end{equation}
where $s_i$ denotes the $i$-th singular value. The cumulative retained energy up to rank $m$ is then defined as
\begin{equation}
c_m = \sum_{i=1}^{m} e_i.
\label{eq:cumulative}
\end{equation}
We select the smallest integer $m$ such that the cumulative retained energy exceeds the threshold $\varepsilon$, where $\varepsilon \in (0,1]$ controls the fraction of preserved spectral energy (see the ablation study in Section~\ref{sec:ablation}). This truncation defines a rank-$m$ principal subspace that preserves the dominant low-rank structure of $\mathbf{F}$ while discarding low-energy components associated with redundancy and noise. 
Empirically, the singular spectrum of visual token representations decays rapidly, confirming that most spectral energy concentrates in a small number of dominant directions and that substantial redundancy exists across the token sequence. Crucially, $\varepsilon$ controls the expressiveness of the retained subspace rather than the compression level, which is controlled independently by the token budget $k$ at the selection stage.



\subsection{Token Contribution via Leverage Scores}
\label{ssec:leverage}

Having identified the dominant subspace, we quantify each token's contribution using leverage scores. For each token $t \in \mathbb{N}$, the rank-$m$ leverage score is defined as:

\begin{equation}
\ell_t = \frac{1}{m} \sum_{j=1}^m (U_{t,j})^2 = \frac{1}{m} \|\mathbf{U}_{t,[1:m]}\|_2^2,
\label{eq:leverage_score}
\end{equation}
where $\ell_t$ represents the average squared projection of token $t$ onto the top-$m$ principal directions. Because the columns of $\mathbf{U}$ are orthonormal, the leverage scores sum to 1 across all tokens ($\sum_t \ell_t = 1$), allowing them to be interpreted as a normalized importance distribution over the token set. Leverage scores naturally quantify how strongly each token participates in the dominant low-rank subspace. Tokens with high $\ell_t$ strongly align with the principal semantic patterns and are therefore highly representative of the global spectral energy, whereas low-leverage tokens mainly reside in the discarded subspace and contribute little to the preserved information.

\subsection{Token Selection and Pruning}
\label{ssec:selection}
Our goal is to identify the minimal subset of tokens that collectively preserves most of the dominant signal. To this end, we first sort the tokens in descending order according to their leverage scores. 
We then retain the top-$k$ tokens, where $k$ corresponds to the desired pruning rate. The $k$ token's indices retained are then re-sorted into their original spatial order to preserve positional embeddings and compatibility with downstream attention mechanisms. This preserves the original positional indices, allowing the retained tokens to reuse their corresponding positional embeddings without modification. Tokens not selected are discarded. The resulting pruned feature matrix $\mathbf{F}' \in \mathbb{R}^{k \times d}$ approximates the original spectral energy structure while substantially reducing sequence length. This process provides a principled spectral-energy-aware pruning strategy that outperforms local heuristics by prioritizing tokens that collectively span the essential low-rank subspace. 

\section{EXPERIMENTS AND ANALYSIS}
\label{sec:experiments_analysis}

\begin{table} [th]
\centering
\caption{Comparison under varying vision token budgets. Best and second-best results within each budget group are marked in \textbf{bold} and \underline{underlined}, respectively.}
\label{tab:performance-comparison}
\small
\setlength{\tabcolsep}{7pt}
 
\begin{tabular}{l c c c c}
\toprule
\textbf{Methods} & \textbf{POPE}  & \textbf{GQA} & \textbf{TextVQA}  & \textbf{MME}\\
\midrule
\rowcolor{gray!20}
Vanilla & 86.96 & 61.90 & 58.20 & 1862.00 \\
\midrule
\multicolumn{5}{c}{\textit{Retain 192 Tokens (33.3\%)}} \\
\midrule
ToMe~\cite{bolya2023tome} & 72.40 & 54.30 & 52.10 & 1563.00 \\
FastV~\cite{chen2024fastv}  & 64.80 & 52.88 & 52.50 & 1605.00 \\
PDrop~\cite{wang2025PDrop}  & \underline{82.30} & \underline{57.30} & \underline{56.50} & \underline{1766.00} \\
Ours &  \textbf{87.75} & \textbf{59.88} & \textbf{57.24} & \textbf{1788.00} \\
\midrule
\multicolumn{5}{c}{\textit{Retain 128 Tokens (22.2\%)}} \\
\midrule
ToMe~\cite{bolya2023tome}  & 62.80 & 52.40 & 49.10 & 1343.00 \\
FastV~\cite{chen2024fastv}  &  53.40 & 49.60 & 50.60 & 1490.00 \\
PDrop~\cite{wang2025PDrop}  &  82.30 & 57.10 & \textbf{56.60} & 1664.00 \\
VisionZip~\cite{yang2025visionzip} & \underline{83.20} & \underline{57.60} & 55.80 & \textbf{1761.70} \\
Ours &  \textbf{86.73} & \textbf{58.70} & \underline{56.14} & \underline{1674.00} \\
\midrule
\multicolumn{5}{c}{\textit{Retain 64 Tokens (11.1\%)}} \\
\midrule
ToMe~\cite{bolya2023tome}  &  52.50 & 48.60 & 45.30 & 1138.00 \\
FastV~\cite{chen2024fastv}  & 38.20 & 46.10 & 47.80 & 1255.00 \\
PDrop~\cite{wang2025PDrop}  & 55.90 & 47.50 & 50.60 & 1092.00 \\
SparseVLM~\cite{zhang2025sparsevlm} &  \underline{77.50} & \underline{53.70} & \underline{53.40} & \underline{1559.00} \\
Ours & \textbf{83.87} & \textbf{53.77} & \textbf{55.14} & \textbf{1575.00} \\
\midrule
\multicolumn{5}{c}{\textit{Retain 32 Tokens (5.6\%)}} \\
\midrule
SparseVLM~\cite{zhang2025sparsevlm} &  67.90 & 48.30 & 46.10 & 1046.70 \\
VisionZip~\cite{yang2025visionzip} &  \underline{68.70} & \underline{51.80} & \underline{53.10} & \underline{1247.40} \\
Ours &  \textbf{79.34} & \textbf{53.52} & \textbf{54.81} & \textbf{1436.00} \\
\midrule
\multicolumn{5}{c}{\textit{Retain 16 Tokens (2.8\%)}} \\
\midrule
Ours &  73.90 & 53.00 & 54.00 & 1281.00 \\
\bottomrule
\end{tabular}

\end{table}

\subsection{Experimental settings}
\label{ssec:experimental_settings}
In this study, we adopt LLaVA-1.5-7B~\cite{liu2023llava} as the baseline model. We evaluate our method on widely used multimodal benchmarks, including GQA~\cite{hudson2019gqa}, TextVQA~\cite{singh2019towards}, POPE~\cite{li2023pope}, and MME~\cite{fu2023mme}, which collectively assess complementary capabilities ranging from compositional visual reasoning and text-centric understanding to scientific reasoning, hallucination robustness, and general multimodal perception.
Following standard LLaVA-1.5 settings, input images are resized to a resolution of $336 \times 336$, resulting in 576 vision tokens. All experiments are conducted using PyTorch on a single NVIDIA RTX 3080 16 GB GPU with an Intel Core i7-11800H CPU. 

\subsection{Comparative Evaluation}
\label{sec:Comparison}
We compare our method against representative encoder-side (ToMe, 
VisionZip, TRIM) and decoder-side (FastV, PyramidDrop, SparseVLM) pruning approaches under varying post-pruning token budgets ranging from 192 to 16 vision tokens. The results are summarized in Table~\ref{tab:performance-comparison}.
SVD-Prune exhibits strong robustness to token pruning rate on GQA. It incurs only a 2.02-point drop at 192 tokens and remains competitive at 128 tokens (-3.20 points), outperforming all prior methods at these budgets. Even under low compression, it degrades gracefully, achieving 53.77 at 64 tokens and maintaining stable performance at 32 and 16 tokens (-8.38 and -8.86 points, respectively). On TextVQA, where accurate text localization and contextual reasoning are essential, our method demonstrates high stability 
It incurs minimal degradation at 192 and 128 tokens (-0.96 and -2.06 points) and achieves the best performance at 64 tokens (-3.06 points). Even at 32 and 16 tokens, performance remains stable (-3.39 and -4.17 points), preserving text-relevant visual information under severe compression.

\subsection{Computational Overhead Analysis}
\label{ssec:computational}

\begin{table} [t]
\centering
\caption{FLOPs breakdown as a function of retained vision tokens.}

\label{tab:flop}
\small
\setlength{\tabcolsep}{5pt}
\begin{tabular}{r c c c c c }
\toprule

Tokens & Encoder & Projector  & Decoder  & Total & Reduction  \\
     &  [G] &  [G] &  [T] &  [T] &  [\%] \\
\midrule
\rowcolor{gray!20}
576  & 190.6 & 12.080 & 3.250 & 3.450 & 0.00 \\
\midrule
192  &  & 4.030 & 1.230 & 1.430 & 58.7 \\
128  &  & 2.680 & 0.903 & 1.100 & 68.2 \\
64   & 190.6 & 1.340 & 0.576 & 0.770 & 77.7 \\
32   &  & 0.671 & 0.413 & 0.604 & 82.5 \\
16   &  & 0.336 & 0.332 & 0.523 & 84.8 \\

\bottomrule
\end{tabular}
\end{table}

Table~\ref{tab:flop} analyzes the computational impact of vision token reduction. While the computational cost of the vision encoder remains constant, the costs of both the projector and the LLM scale linearly with the number of retained vision tokens. Consequently, aggressive vision token pruning yields substantial efficiency gains, reducing total FLOPs by 58.7\%, 68.2\%, 77.7\%, 82.5\%, and up to 84.8\% when retaining 192, 128, 64, 32, and 16 tokens, respectively. In particular, reducing the token count from 576 to 16 lowers total computation from 3.45\,T to 0.52\,T FLOPs, highlighting that vision token count is the primary driver of inference cost.

\subsection{ Efficiency Analysis on Commodity GPU Hardware}

Table~\ref{tab:efficiency} demonstrates the practical efficiency gains achieved by SVD-Prune on commodity GPU hardware. Using only 16 visual tokens instead of 576 significantly lowers inference latency from 930.00 ms to 222.56 ms, corresponding to a \(4.18\times\) speedup and an increase in throughput from 1.06 to 4.33 FPS. 
Kernel execution time is also reduced by nearly \(2\times\), confirming that pruning redundant visual tokens effectively alleviates the computational burden of multimodal processing. 
Notably, peak VRAM usage decreases only modestly despite the aggressive token pruning. This suggests that memory consumption is still largely dominated by the language model parameters and intermediate activations rather than by visual tokens alone.

\begin{table}[ht]
\centering
\caption{Efficiency analysis on the GQA dataset.
}
\label{tab:efficiency}

\small
\setlength{\tabcolsep}{3pt}

\begin{tabular}{@{}lccccc@{}}
\toprule
Method & Tokens & Latency & VRAM & FPS & Kernel time \\
       &        & [ms]    & [GB] &     & [ms] \\
\midrule
Vanilla  & 576 & 930.00 & 14.11 & 1.06 & 670.03 \\
SVD-Prune & 16  & 222.56 & 13.82 & 4.33 & 338.34 \\
\midrule
\midrule
 & $\times$36.0$\downarrow$ & $\times$4.18$\downarrow$ & $-$0.29 & $\times$4.08$\uparrow$ & $\times$1.98$\uparrow$ \\
\bottomrule
\end{tabular}

\end{table}


\begin{table}[th]
\centering
\small
\setlength{\tabcolsep}{14pt}
\caption{
SVD-Prune overhead analysis on the GQA dataset for the 16-token setting.} 
\label{tab:latency_breakdown}
\begin{tabular}{l c c c}
\toprule
\textbf{Stage} & \textbf{Vanilla} & \textbf{Pruned}  \\
\midrule
Encoder & $22.88$     & $22.88$  \\
\midrule

\multicolumn{3}{c}{\textbf{SVD-Prune}} \\
\midrule
 SVD Decomposition           & N/A                    & $23.94 $    \\
Leverage Score   & N/A                    & $0.53$      \\
Token Selection             & N/A                    & $0.26 $    \\
\midrule
Decoder  & $907.12$   & $174.95$  \\
\bottomrule
\end{tabular}

\end{table}

Importantly, the computational overhead of SVD-Prune remains negligible. As shown in Table~\ref{tab:latency_breakdown}, the pruning stage---comprising SVD decomposition (23.94\,ms), leverage score computation (0.53\,ms), and token selection (0.26\,ms)---adds only 24.73\,ms, compared to the 732.17\,ms saved during decoding.

\subsection[Ablation study on the threshold epsilon]{Ablation study on the threshold $\varepsilon$}
\label{sec:ablation}

\begin{figure} [th] 
\centering
\includegraphics[width=1.0\linewidth]{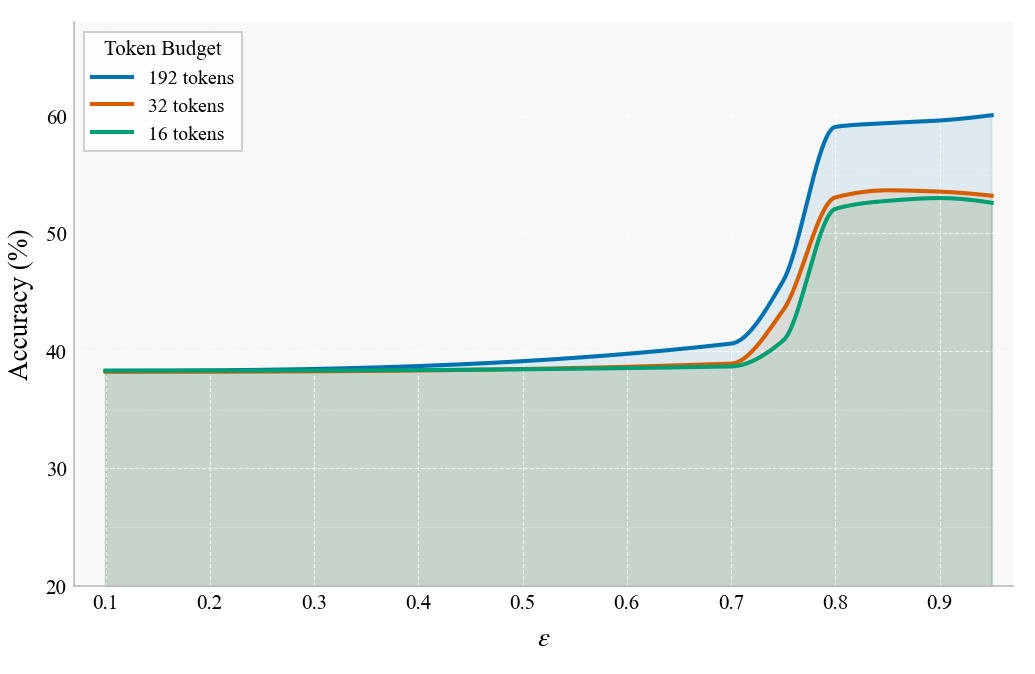}
\caption{Impact of the threshold $ \varepsilon $ on accuracy for three token budgets: 192, 32, and 16 tokens.}

\label{fig:epsilon_ablation}
\end{figure}

To find the optimal threshold $ \varepsilon $, we perform an ablation study on GQA dataset across three token budgets (16, 32, and 192 tokens). As shown in Fig.~\ref{fig:epsilon_ablation}, accuracy remains nearly flat at approximately 38--39\% for $ \varepsilon \in [0.1, 0.7] $, confirming that low-sepctral singular directions carry negligible discriminative information and can be safely discarded. Beyond $ \varepsilon = 0.7$, accuracy rises sharply and consistently across all budgets, reaching 59.9\%, 53.5\%, and 53.0\% for the 192, 32, and 16-token regimes respectively at $ \varepsilon = 0.8 $. This demonstrates that retaining dominant spectral energy components is critical for preserving task-relevant visual content. Setting $ \varepsilon \in$ [0.8, 0.9] yields the best accuracy-efficiency tradeoff, with diminishing returns beyond 0.9 and a slight decrease at $ \varepsilon = 0.95 $  for the 32 and 16-token regimes, suggesting mild sensitivity to noise at very high energy retention. We therefore set $ \varepsilon = 0.8 $ as our default threshold, corresponding to the inflection point of the accuracy curve.

\section{Conclusion}
\label{conclusion}
In this work, we introduced SVD-Prune, a training-free vision token pruning method based on global low-rank structure analysis. Unlike local scoring heuristics, SVD provides a global characterization of visual representations by jointly analyzing all tokens, enabling the identification of shared informative structures across the image (e.g., edges, textures, objects). 
SVD-Prune consistently outperforms existing pruning methods, particularly in the low-token regime. 
At 16 retained tokens, it reduces total FLOPs by up to 84.8\% and achieves a 4.18$\times$ latency speedup while maintaining strong multimodal performance. These results demonstrate that substantial vision token reduction is possible without heavily compromising downstream reasoning capabilities, making efficient vision--language inference feasible under strict computational constraints. More broadly, our findings highlight the promise of spectral, globally informed token selection as an alternative to attention-based or local pruning strategies.\\






\textbf{Acknowledgments}\\
This work was supported by the European Union's Horizon Europe research and innovation programme under the Marie Skłodowska-Curie COFUND grant agreement No 101127936 (DeMythif.AI), and France 2030 funding managed by the National Research Agency (ANR), as part of IA CLUSTER program, reference ANR-23-IACL-0003-DATAIA CLUSTER.




\end{document}